# Evolutionary Turing in the Context of Evolutionary Machines

**Mark Burgin**

Dept. of Mathematics

University of California

405 Hilgard Avenue

Los Angeles, CA 90095, USA

**Eugene Eberbach**

Dept. of Eng. and Science

Rensselaer Polytechnic Institute

275 Windsor Street

Hartford, CT 06120, USA

**Abstract**: One of the roots of evolutionary computation was the idea of Turing about unorganized machines. The goal of this work is the development of foundations for evolutionary computations, connecting Turing's ideas and the contemporary state of art in evolutionary computations. To achieve this goal, we develop a general approach to evolutionary processes in the computational context, building mathematical models of computational systems, functioning of which is based on evolutionary processes, and studying properties of such systems. Operations with evolutionary machines are described and it is explored when definite classes of evolutionary machines are closed with respect to basic operations with these machines. We also study such properties as linguistic and functional equivalence of evolutionary machines and their classes, as well as computational power of evolutionary machines and their classes, comparing of evolutionary machines to conventional automata, such as finite automata or Turing machines.

# 1. Introduction

Alan Turing had many other ideas in computer science. In particular, Turing (1948) proposed to use what is now called genetic algorithms in his unorganized machines. Turing, while at Cambridge, developed his *automatic machines* (now known as *Turing machines*) and *choice machines*. In 1939, he defended his Ph.D. on *oracle machines* under Alonzo Church supervision at Princeton and during World War II worked on Colossus to break Enigma code. After the end of the war, Turing joined the National Physical Laboratory in 1945 and worked under the supervision of Sir Charles Darwin, producing (in 1948) a report, which can be viewed as a blueprint for the future field of evolutionary computation. Titled Intelligent Machinery, this report was left unpublished until 1968, because Darwin, his boss, considered it to be a "schoolboy essay" not suitable for publication.

In this report, Turing proposed new models of computation, which he called *unorganized machines* (*u-machines*). There were two types of u-machines: based on Boolean networks and based on finite state machines.

- A-*type* and B-*type* u-machines were Boolean networks made up of a fixed number of two-input NAND gates (neurons) and synchronized by a global clock. While in A-type u-machines the connections between neurons were fixed, B-type u-machines had modifiable switch type interconnections. Starting from the initial random configuration and applying a kind of genetic algorithm, B-type u-machines were supposed to learn which of their connections should be on and which off.

- P-*type* u-machines were tapeless Turing machines reduced to their finite state machine control, with an incomplete transition table, and two input lines for interaction: the pleasure and the pain signals.

In his B-type u-machines, Turing pioneered two areas at the same time: neural networks and evolutionary computation (more precisely, evolutionary artificial neural networks), while his P-type u-machines represent reinforcement learning. However, this work had no impact on these fields, due to the unfortunate combination of Turing's death and the twenty-year delay in publication.

Turing was convinced that his B-type u-machine can simulate his universal machine, though he never provided a formal proof. To simulate the infinite tape of a Turing machine, a u-machine with an infinite number of neurons would be needed. This is due to the discrete nature of the neurons, which were based on two input Boolean NAND gates. By contrast, two real-valued neurons are sufficient to model a Turing machine.

B-type u-machines were defined to have a finite number of neurons, and it is not clear whether Turing was aware that infinitely many neurons were needed for the simulation. This inconsistency would certainly have been uncovered when working on the formal proof. But perhaps Turing was aware of it, and expected to have no problems extending his definitions to the infinite case.

In any case, these ideas are one of the roots of evolutionary computation in general and evolutionary computation theory, in particular. Evolutionary computation theory is still very young and incomplete. Until recently, evolutionary computation did not have a theoretical model that represented practice in this domain. Very little has been known about expressiveness, or computational power, of evolutionary computation (EC) and its scalability. Of course, there are many results on the theory of evolutionary algorithms (see, e.g., [10, 11, 13, 15]). Theoretical topics studied in evolutionary computations include convergence in the limit (elitist selection, Michalewicz's contractive mapping genetic algorithms, ((1+1)-ES), convergence rate (Rechenberg's 1/5 rule), the Building Block analysis (Schema Theorems for GA and GP), best variation operators (No Free Lunch Theorem). However, these authors do not introduce automaton models – rather they apply high-quality mathematical apparatus to existing process models, such as Markov chains, etc. They also cover only some aspects of evolutionary computation like convergence or convergence rate, neglecting for example evolutionary computation expressiveness, self-adaptation, or scalability. In other words, evolutionary computation is not treated as a distinct and complete area with its own distinct model situated in the context of general computational models. This means that in spite of intensive usage of mathematical techniques, theoretical foundations of evolutionary computations are only on the first stage of creation. As a result, many properties of evolutionary processes could not be precisely studied or even found by researchers. Our research is aimed at filling this gap by building and developing further rigorous mathematical foundations of evolutionary computations.

In [8], the evolutionary Turing machine model was proposed to provide more rigorous foundations for evolutionary computation. As it is proved in [4], an evolutionary Turing machine is an extension of the conventional Turing machine, going beyond the Turing machine as an important type of super-recursive algorithms [2]. In several papers, the authors introduced and studied more general and more powerful than evolutionary Turing machines models to reflect cooperation and competition [4], self-evolution [9], universality [5], and expressiveness of evolutionary finite automata [6]. The most general model of evolutionary computations is evolutionary automaton/machine (EA). There two general types of evolutionary automata/machines – basic evolutionary automata/machines and general evolutionary automata/machines. All other classes of evolutionary automata/machines, such as evolutionary finite automata, evolutionary Turing machines or evolutionary inductive Turing machines, are special cases of one of these two types.

In this work, we develop a general approach to evolutionary processes in the computational context, build mathematical models of the systems functioning of which is based on evolutionary processes and study properties of such systems. Two classes are introduced: basic evolutionary automata/machines and general evolutionary automata/machines. Relations between computing power of these classes are explored using operations with evolutionary automata/machines. We also consider such properties as linguistic and functional equivalence of evolutionary automata/machines and their classes.

## 2. Modeling Evolution by Evolutionary Machines

Evolutionary computations are artificial intelligence processes based on natural selection and evolution. Evolutionary computations are directed by evolutionary algorithms. In technical terms, an evolutionary algorithm is a probabilistic beam hill climbing search algorithm directed by the chosen fitness function. To formalize this concept in mathematically rigorous terms, we define a formal algorithmic model of evolutionary computation - an *evolutionary automaton*, which is also called an *evolutionary machine*.

Let **K** be a class of automata.

**Definition 2.1.** A *basic evolutionary* **K**-*machine* (BEM), also called *basic evolutionary* **K**-*automaton*, is a (possibly infinite) sequence $E = \{A[t]; t = 0, 1, 2, 3, ... \}$ of automata $A[t]$ from **K** each working on the population/generation $X[t]$ ($t = 0, 1, 2, 3, ...$) where:

- the automaton $A[t]$ called a *component*, or more exactly, a *level automaton*, of $E$ represents (encodes) a one-level evolutionary algorithm that works with the population/generation *X*[*t*] of the population by applying the variation operators *v* and selection operator *s*;

- the first population/generation *X*[0] is given as input to *E* and is processed by the automaton *A*[0], which generates/produces the first population/generation *X*[1] as its output, which goes to the automaton *A*[1];

- for all $t = 1, 2, 3, \ldots$, the population/generation *X*[*t* + 1] is obtained by applying the variation operator *v* and selection operator *s* to the population/generation *X*[*t*] and these operations are performed by the automaton *A*[*t*], which receives *X*[*t*] as its input;

- the goal of the BEM *E* is to build a population *Z* satisfying the search condition.

The desirable search condition is the optimum of the fitness performance measure *f*(*x*[*t*]) of the best individual from the population/generation *X*[*t*]. There are different modes of the EM functioning and different termination strategies. When the search condition is satisfied, then working in the recursive mode, the EM *E* halts (*t* stops to be incremented), otherwise a new input population/generation *X*[*t* + 1] is generated by the automaton *A*[*t*]. In the inductive mode, it is not necessary to halt to give the result (cf. [5]). When the search condition is satisfied and *E* is working in the inductive mode, the EM *E* stabilizes (the population/generation *X*[*t*] stops changing), otherwise a new input population/generation *X*[*t* + 1] is generated by A[*t*].

We denote the class of all basic evolutionary machines with level automata from **K** by **BEAK**.

**Definition 2.2.** A *general evolutionary* **K**-*machine* (GEM), also called *general evolutionary* **K**-*automaton*, is a (possibly infinite) sequence $E = \{A[t]; t = 0, 1, 2, 3, \ldots\}$ of automata *A*[*t*] from **K** each working on population/generation *X*[*i*] where:

- the automaton *A*[*t*] called a *component*, or more exactly, a *level automaton*, of *E* represents (encodes) a one-level evolutionary algorithm that works with populations/generations *X*[*i*] of the population by applying the variation operators *v* and selection operator *s*;

- the first population/generation *X*[0] is given as input to *E* and is processed by the automaton *A*[0], which generates/produces the first population/generation *X*[1] as its output, which goes to the automaton *A*[1];

- for all $t = 1, 2, 3, \ldots$, the automaton *A*[*t*], which receives *X*[*i*] as its input either from *A*[*t* + 1] or from *A*[*t* - 1], then *A*[*t*] applies the variation operator *v* and selection operator *s* to the population/generation *X*[*t*], producing the population/generation *X*[*t* + 1] and sending this generation either to *A*[*t* + 1] or to *A*[*t* - 1];

- the goal of the GEM *E* is to build a population *Z* satisfying the search condition.

We denote the class of all general evolutionary **K**-machines **GEAK**. As any basic evolutionary **K**-machine is also a general evolutionary **K**-machine, we have inclusion of classes **BEAK** $\subseteq$ **GEAK**.

Let us consider some examples of evolutionary **K**-machines. An important class of evolutionary machines is evolutionary finite automata [6]. Here **K** consists of finite automata.

**Definition 2.3.** A *basic* (*general*) *evolutionary finite automaton* (*EFA*) is a basic (general) evolutionary machine *E* in which all automata *A*[*t*] are finite automata *G*[*t*] each working on the population *X*[*t*] in populations/generations *t* = 0, 1, 2, 3, ... .

We denote the class of all general evolutionary finite automata by **GEFA**.

It is possible to take as **K** deterministic finite automata, which form the class **DFA**, or nondeterministic finite automata, which form the class **NFA**. This gives us four classes of evolutionary finite automata: **BEDFA** (**GEDFA**) of all deterministic basic (general) evolutionary finite automata and **BENFA** (**GENFA**) of all nondeterministic basic (general) evolutionary finite automata.

Evolutionary Turing machines [4, 8] are another important class of evolutionary machines.

**Definition 2.4.** A *basic* (*general*) *evolutionary Turing machine* (*ETM*) *E* = { T[*t*]; *t* = 0, 1, 2, 3, ... } is a basic (general) evolutionary machine *E* in which all automata *A*[*t*] are Turing machines T[*t*] each working on population/generation *X*[*t*] with the generation parameter *t* = 0, 1, 2, 3, ...

Turing machines T[*t*] as components of *E* perform multiple computations [1]. Variation and selection operators are recursive to allow performing level computation on Turing machines.

**Definition 2.5.** A *basic* (*general*) *evolutionary inductive Turing machine* (*EITM*) *EI* = {ITM[*t*]; *t* = 0, 1, 2, 3, ... } is a basic (general) evolutionary machine *E* in which all automata *A*[*t*] are inductive Turing machines ITM[*t*] [2] each working on the population/generation *X*[*t*] with the generation parameter *t* = 0, 1, 2, 3, ...

Simple inductive Turing machines are abstract automata (models of algorithms) closest to Turing machines [2]. The difference between them is that a Turing machine always gives the final result after a finite number of steps and after this it stops or, at least, informs when the result is obtained. Inductive Turing machines also give the final result after a finite number of steps, but in contrast to Turing machines, inductive Turing machines do not always stop the process of computation or inform when the final result is obtained. In some cases, they do this, while in other cases they continue their computation and give the final result. Namely, when the content of the output tape of a simple inductive Turing machine forever stops changing, it is the final result.

**Definition 2.6.** A basic (general) evolutionary inductive Turing machine (EITM) $EI = \{ITM[t]; t = 0, 1, 2, 3, ... \}$ has order $n$ if all inductive Turing machines $ITM[t]$ have order less than or equal to $n$ and at least, one inductive Turing machine $ITM[t]$ has order $n$.

We remind that inductive Turing machines with recursive memory are called *inductive Turing machines of the first order* [2]. The memory $E$ is called *n-inductive* if its structure is constructed by an inductive Turing machine of order $n$. Inductive Turing machines with *n*-inductive memory are called *inductive Turing machines of order n + 1*.

We denote the class of all evolutionary inductive Turing machines of order $n$ by **EITM**$_n$.

**Definition 2.7.** A *basic* (*general*) *evolutionary limit Turing machine* (*ELTM*) $EI = \{LTM[t]; t = 0, 1, 2, 3, ... \}$ is a basic (general) evolutionary machine $E$ in which all automata $A[t]$ are limit Turing machines $LTM[t]$ [2] each working on the population/generation $X[t]$ in with the generation parameter $t = 0, 1, 2, 3, ...$

When the search condition is satisfied, then the evolutionary limit Turing machine $EI$ stabilizes (the population $X[t]$ stops changing), otherwise a new input population $X[t + 1]$ is generated by $LTM[t]$.

We denote the class of all evolutionary limit Turing machines of the first order by **ELTM**.

Basic and general evolutionary **K**-machines from **BEAK** and **GEAK** are called *unrestricted* because sequences of the level automata A[t] and the mode of the evolutionary machines functioning are arbitrary. For instance, there are *unrestricted evolutionary Turing machines* when **K** is equal to **T** and *unrestricted evolutionary finite automata* when **K** is equal to **FA**.

However it is possible to consider only basic (general) evolutionary **K**-machines from **BEAK** (**GEAK**) in which sequences of the level automata have some definite type $Q$. Such machines are called *Q-formed basic* (*general*) *evolutionary* **K**-*machines* and their class is denoted by **BEAK**$^Q$ for basic machines and **GEAK**$^Q$ for general machines.

When the type **Q** contains all finite sequences, we have *bounded* basic (general) evolutionary **K**-machines. Some classes of bounded basic evolutionary **K**-machines are studied in [7] for such classes **K** as finite automata, push down automata, Turing machines, or inductive Turing machines, i.e., such classes as bounded basic evolutionary Turing machines or bounded basic evolutionary finite automata

When the type **Q** contains all periodic sequences, we have *periodic* basic (general) evolutionary **K**-machines. Some classes of periodic basic evolutionary **K**-machines are studied in [7] for such classes **K** as finite automata, push down automata, Turing machines, inductive Turing machines and limit Turing machines. Note that while in a general case, evolutionary automata cannot be codified by finite words, periodic evolutionary automata that can be codified by finite words.

Another condition on evolutionary machines determines their mode of functioning or computation. Here we consider the following modes of functioning/computation.

1. The *finite-state mode*: any computation is going by state transition where states belong to a fixed finite set.

2. The *bounded mode*: the number of generations produced in all computations is bounded by the same number.

3. The *terminal* or *finite mode*: the number of generations produced in any computation is finite.

4. The *recursive mode*: in the process of computation, it is possible to reverse the direction of computation, i.e., it is possible to go from higher levels to lower levels of the automaton, and the result is defined after finite number of steps.

5. The *inductive mode*: the computation goes into one direction, i.e., without reversions, and if for some $t$, the generation $X[t]$ stops changing, i.e., $X[t] = X[q]$ for all $q > t$, then $X[t]$ is the result of computation.

6. The *inductive mode with recursion*: recursion (reversion) is permissible and if for some $t$, the generation $X[t]$ stops changing, i.e., $X[t] = X[q]$ for all $q > t$, then $X[t]$ is the result of computation.

7. The *limit mode*: the computation goes into one direction and the result of computation is the limit of the generations $X[t]$.

8. The *limit mode with recursion*: recursion (reversion) is permissible and the result of computation is the limit of the generations $X[t]$.

These modes determine the type of functioning for any (not only evolutionary) automata and they are complementary to the three traditional modes of computing automata: computation, acceptation and decision/selection [3].

Existence of different modes of computation shows that the same algorithmic structure of an evolutionary automaton/machine *E* provides for different types of evolutionary computations.

We see that only general evolutionary machines allow recursion. In basic evolutionary machines, the process of evolution (computation) goes strictly in one direction. Thus, general evolutionary machines have more possibilities than basic evolutionary machines and it is interesting to relations between these types of evolutionary machines. This is done in the next section.

Note that utilization of recursive steps in evolutionary machines provides means for modeling reversible evolution, as well as evolution that includes periods of decline and regression.

### 3. Computing and accepting power of evolutionary machines

As we know from the theory of automata and computation, it is proved that different automata or different classes of automata are equivalent. However there are different kinds of equivalence. Here we consider two of them: functional equivalence and linguistic equivalence.

**Definition 3.1** [3]**.** a) Two automata *A* and *B* are *functionally equivalent* if given the same input, they give the same output. b) Two classes of automata *A* and *B* are *functionally equivalent* if for any automaton from *A*, there is a functionally equivalent automaton from *B* and vice versa.

Functional equivalence of automata means that these automata can compute the same functions. Functional equivalence of classes of automata means that the same class of functions is computable by both classes.

For instance, it is proved that deterministic and nondeterministic Turing machines are functionally equivalent [12]. Similar results are true for evolutionary finite automata and conventional finite automata.

**Theorem 3.1** [7]. For any basic *n*-level evolutionary finite automaton *E*, there is a finite automaton $A_E$ functionally equivalent to *E*.

Here we study relations between basic and general evolutionary machines, assuming that all these machines work in the terminal mode. Let *P* be a function such that $P(x, i) = i$ for any *x* and *i*.

**Definition 3.2** [3]. The *P-conjunctive parallel composition* $\wedge_P A_i$ of the algorithms/automata $A_i$ ($i = 1, 2, 3, \ldots, n$) is an algorithm/automaton *D* such that the result of application of *D* to any input *u* is equal to $A_i(u)$ when $P(u) = i$.

This concept allows us to show in a general case of the terminal mode that basic and general evolutionary machines are equivalent.

**Theorem 3.2.** If a class **K** is closed with respect to *P*-conjunctive parallel composition, then for any general evolutionary **K**-machine, there is a functionally equivalent basic evolutionary **K**-machine.

**Corollary 3.1.** If a class **K** is closed with respect to *P*-conjunctive parallel composition, then classes **GEAK** and **BEAK** are functionally equivalent.

The class **T** of all Turing machines is closed with respect to *P*-conjunctive parallel composition [3]. Thus, Theorem 3.2 implies the following result.

**Corollary 3.2.** Classes **GEAT** of all general evolutionary Turing machines and **BEAT** of all basic evolutionary Turing machines are functionally equivalent.

The class **IT** of all inductive Turing machines is closed with respect to *P*-conjunctive parallel composition [3]. Thus, Theorem 3.2 implies the following result.

**Corollary 3.3.** Classes **GEAIT** of all general evolutionary inductive Turing machines and **BEAIT** of all basic evolutionary inductive Turing machines are functionally equivalent.

**Corollary 3.4.** Classes $\textbf{GEAIT}_n$ of all general evolutionary inductive Turing machines of order *n* and $\textbf{BEAIT}_n$ of all basic evolutionary inductive Turing machines of order *n* are functionally equivalent.

The same is true for evolutionary limit Turing machines.

**Corollary 3.5.** Classes **GEALT** of all general evolutionary limit Turing machines and **BEALT** of all basic evolutionary limit Turing machines are functionally equivalent.

**Definition 3.3** [3]**.** a) Two automata *A* and *B* are *linguistically equivalent* if they accept (generate) the same language.

b) Two classes of automata *A* and *B* are *linguistically equivalent* if they accept (generate) the same class of languages.

For instance, it is proved that deterministic and nondeterministic finite automata are linguistically equivalent or that deterministic and nondeterministic Turing machines are linguistically equivalent [12].

It is proved in [3] that functional equivalence is stronger than linguistic equivalence because two functionally equivalent automata are always linguistically equivalent. This allows us to obtain the following results using the operation with abstract automata called *P*-conjunctive parallel composition, which is described in [3].

**Theorem 3.3.** If a class **K** is closed with respect to *P*-conjunctive parallel composition, then for any general evolutionary **K**-machine, there is a linguistically equivalent basic evolutionary **K**-machine.

This result shows how closure properties influence computational power of evolutionary machines.

**Corollary 3.6.** If a class **K** is closed with respect to *P*-conjunctive parallel composition, then classes **GEAK** and **BEAK** are linguistically equivalent.

The class **T** of all Turing machines is closed with respect to *P*-conjunctive parallel composition [5]. Thus, Theorem 3.3 implies the following result.

**Corollary 3.7.** Classes **GEAT** of all general evolutionary Turing machines and **BEAT** of all basic evolutionary Turing machines are linguistically equivalent.

The class **IT** of all inductive Turing machines is closed with respect to *P*-conjunctive parallel composition [3]. Thus, Theorem 3.3 implies the following result.

**Corollary 3.8.** Classes **GEAIT** of all general evolutionary inductive Turing machines and **BEAIT** of all basic evolutionary inductive Turing machines are linguistically equivalent.

**Corollary 3.9.** Classes **GEAIT**$_n$ of all general evolutionary inductive Turing machines of order *n* and **BEAIT**$_n$ of all basic evolutionary inductive Turing machines of order *n* are linguistically equivalent.

The same is true for evolutionary limit Turing machines.

**Corollary 3.10.** Classes **GEALT** of all general evolutionary limit Turing machines and **BEALT** of all basic evolutionary limit Turing machines are linguistically equivalent.

Obtained results allow us to solve the following problem formulated in [7].

**Problem 1** [7]. Are periodic evolutionary finite automata more powerful than finite automata?

To solve it, we need additional properties of periodic evolutionary finite automata.

**Theorem 3.4.** Any general (basic) periodic evolutionary finite automaton $F$ with the period $k > 1$ is functionally equivalent to a periodic evolutionary finite automaton $E$ with the period 1.

Proof. For basic periodic evolutionary finite automata, this result is proved in [7]. The proof for general periodic evolutionary finite automata is similar based on the fact that the class of finite automata is closed with respect to sequential composition and iteration [12].

**Corollary 3.11.** Any general (basic) periodic evolutionary finite automaton $F$ with the period $k > 1$ is linguistically equivalent to a periodic evolutionary finite automaton $E$ with the period 1.

**Theorem 3.5.** Any basic periodic evolutionary finite automaton $F$ is linguistically equivalent to a finite automaton.

**Corollary 3.12.** Basic periodic evolutionary finite automata have the same accepting power as finite automata.

The next result demonstrates functional equivalence between evolutionary machines and conventional automata.

**Theorem 3.6.** Any general periodic evolutionary finite automaton $E$ is functionally equivalent to a one-dimensional cellular automaton.

As one-dimensional cellular automata are equivalent to Turing machines [2], we have the following result.

**Corollary 3.13.** General periodic evolutionary finite automata have the same accepting power as Turing machines.

Consequently, we have the following result because by Theorem 3.5, basic periodic evolutionary finite automata are linguistically equivalent to finite automata.

**Corollary 3.14.** General periodic evolutionary finite automata have more accepting power than basic periodic evolutionary finite automata and than finite automata.

These results also allow us to solve Problem 4 from [7].

**Problem 4** [7]. What class of languages is generated/accepted by periodic evolutionary finite automata?

Namely, we have the following results.

**Corollary 3.15.** The class of languages is generated/accepted by basic periodic evolutionary finite automata coincides with regular languages.

**Corollary 3.16.** The class of languages is generated/accepted by general periodic evolutionary finite automata coincides with recursively enumerable languages.

Note that for unrestricted evolutionary finite automata results of Theorems 3.5, 3.6 and their corollaries are not true. Namely, we have the following result.

**Theorem 3.7.** The class **GEAFA** of general unrestricted evolutionary finite automata and the class **BEAFA** of basic unrestricted evolutionary finite automata have the same accepting power.

This result means that in the case of evolutionary finite automata, the possibility to move processed data in two directions instead of one direction of the basic evolutionary finite automata does add computing power to these evolutionary automata.

**Conclusion**

We introduced two fundamental classes of evolutionary machines/automata: general evolutionary machines and basic evolutionary machines, exploring relations between these classes. Problems of generation of evolutionary machines/automata by automata from a given class are also studied. Examples of such evolutionary machines are evolutionary Turing machines generated by Turing machines and evolutionary inductive Turing machines generated by inductive Turing machines.

There are open problems about relative computational power of evolutionary automata/machines and conventional abstract automata important for the development of foundations of evolutionary computations.

**Problem 1.** Can an inductive Turing machine of the first order simulate an arbitrary periodic evolutionary inductive Turing machine of the first order?

This problem has a more general form.

**Problem 2.** Can an inductive Turing machine of order $n$ simulate an arbitrary periodic evolutionary inductive Turing machine of order $n$?

The next two problems are weaker cases of this problem.

**Problem 3.** Can an inductive Turing machine of order $n$ simulate an arbitrary periodic evolutionary inductive Turing machine of order $n - 1$?

**Problem 4.** Can an inductive Turing machine of order *n* simulate an arbitrary periodic evolutionary inductive Turing machine of order 1?

Another groups of problems is related to the computational power of two basic types of evolutionary automata/machines

**Problem 5.** Are there necessary and sufficient conditions for general evolutionary machines to be more powerful than basic evolutionary machines?

As we can see from results of this paper, in some cases general evolutionary machines are more powerful than basic evolutionary machines, e.g., for all evolutionary finite automata, while in other cases, it is not true, e.g., for all periodic evolutionary machines.

**Problem 6.** Is it possible for basic evolutionary machines to be more powerful than general evolutionary machines?

There is also a related problem.

**Problem 7.** Under what conditions basic evolutionary machines and general evolutionary machines have the same power?